\documentclass[conference]{IEEEtran}
\IEEEoverridecommandlockouts

\usepackage{cite}
\usepackage{amsmath,amssymb,amsfonts}
\usepackage{algorithmic}
\usepackage{graphicx}
\usepackage{textcomp}
\usepackage{xcolor}
\usepackage{url}
\usepackage{multirow}
\usepackage{amsfonts}
\usepackage{pifont}

\def\BibTeX{{\rm B\kern-.05em{\sc i\kern-.025em b}\kern-.08em
    T\kern-.1667em\lower.7ex\hbox{E}\kern-.125emX}}
\begin{document}
\title{H3DE-Net: Efficient and Accurate 3D Landmark Detection in Medical Imaging\\
\thanks{This work was supported by: (1) National Social Science Foundation in 2024 (No. 24BMZ101); (2) 2023 Project of the 14th Five-Year Plan for Scientific Research of the State Language Commission (No. YB145-73); (3) Jilin Provincial Science and Technology Development Program Project (No. 20250203109SF); and (4) Changchun Humanities and Sciences College Project (No. 2025KG29).}
}

\author{
\IEEEauthorblockN{1\textsuperscript{st} Zhen Huang\textsuperscript{\dag}}
\IEEEauthorblockA{\textit{School of Computer Science}\\ USTC \\
Hefei, Anhui, China}
\and
\IEEEauthorblockN{2\textsuperscript{nd} Tao Tang\textsuperscript{\dag}}
\IEEEauthorblockA{\textit{School of Computer Science and Engineering} \\ UESTC \\
Chengdu, Sichuan, China}
\and
\IEEEauthorblockN{3\textsuperscript{rd} Ronghao Xu\textsuperscript{\dag}}
\IEEEauthorblockA{\textit{School of Computer Science} \\ USTC \\
Hefei, Anhui, China}
\and
\IEEEauthorblockN{4\textsuperscript{th} Yangbo Wei\textsuperscript{\dag}}
\IEEEauthorblockA{\textit{Shanghai Jiao Tong University}\\
Shanghai, China}
\and
\IEEEauthorblockN{5\textsuperscript{th} Wenkai Yang}
\IEEEauthorblockA{\textit{ShanghaiTech University}\\
Shanghai, China}
\and
\IEEEauthorblockN{6\textsuperscript{th} Suhua Wang\textsuperscript{*}}
\IEEEauthorblockA{\textit{Changchun Humanities and Sciences College}\\
wangsuhua@ccrw.edu.cn}
\and
\IEEEauthorblockN{7\textsuperscript{th} Xiaoxin Sun}
\IEEEauthorblockA{\textit{School of Information Science, NENU}\\
Changchun, China}
\and
\IEEEauthorblockN{8\textsuperscript{th} Han Li\textsuperscript{*}}
\IEEEauthorblockA{\textit{Technical University of Munich (TUM)}\\
\textit{Munich Center for Machine Learning (MCML)}\\
tum\_han.li@tum.de}

\and
\IEEEauthorblockN{9\textsuperscript{th} Qingsong Yao\textsuperscript{*}}
\IEEEauthorblockA{\textit{Stanford University}\\
Stanford, CA, USA\\
yaoqingsong19@mails.ucas.edu.cn}

\thanks{\textsuperscript{\dag}\,These authors contributed equally to this work.}
\thanks{\textsuperscript{*}\,Corresponding authors.}
}

\maketitle

\begin{abstract}
Landmark detection is essential in medical image analysis, aiding tasks like surgical navigation, diagnosis, and treatment planning. However, it remains challenging due to the need for fine-grained local detail and long-range spatial dependency modeling in high-dimensional volumetric data. Existing approaches struggle to balance accuracy, efficiency, and robustness, especially in cases of sparse landmark distribution, anatomical variability, and noisy or incomplete scans.

We propose \textbf{H3DE-Net}, a hybrid framework combining CNNs for local feature extraction and a lightweight transformer-based attention module for global context. A volumetric bi-level routing attention mechanism reduces computational overhead while preserving long-range dependencies, and multi-scale feature fusion enhances precision and robustness. This design integrates both global and local representations, overcoming the limitations of CNN-only or transformer-only models.

Extensive experiments on a public CT dataset show that H3DE-Net achieves state-of-the-art performance, significantly improving mean radial error (MRE) and success detection rate (SDR) compared to existing methods. The model is robust in challenging scenarios with missing landmarks or anatomical variations, demonstrating its applicability in real-world clinical settings. All code, pretrained weights, and data processing scripts are publicly available for reproducibility and further research.

\end{abstract}


\section{Introduction}

3D landmark detection is a vital task in medical image analysis, widely applied in surgical navigation, disease diagnosis, and treatment planning~\cite{treatment}. Accurate localization of anatomical landmarks provides reliable spatial references for clinicians, enabling automated diagnosis, personalized treatments, and supporting tasks such as image registration, segmentation, and 3D reconstruction~\cite{3Dreconstruction}. These advancements significantly improve clinical decision-making and treatment outcomes. However, automated 3D landmark detection remains highly challenging due to complex anatomical structures, patient-specific variability, low signal-to-noise ratio (SNR), and differences in image resolution~\cite{ huang2024pele,huang2025medatlas, huang2025embedding}.

Traditional landmark detection methods are broadly categorized into regression-based and heatmap-based approaches~\cite{huang2025casemark}. Regression-based methods predict landmark coordinates efficiently but struggle with global context and large-scale variations, especially in complex anatomical scenarios. Heatmap-based methods preserve spatial structures but are sensitive to local variations, occlusions, and low-resolution conditions. Both approaches face limitations in achieving sub-voxel precision and utilizing multi-scale features effectively.

Convolutional neural networks (CNNs) have become a key tool for 3D landmark detection due to their ability to extract hierarchical, multi-scale features from complex medical images~\cite{zhou2024hybrid}. However, CNNs have difficulty modeling long-range dependencies and global context~\cite{SCN}, especially in tasks with sparse or widely distributed landmarks. Transformers, such as Vision Transformers (ViT) and Swin Transformers, excel at modeling global dependencies~\cite{zheng2024smaformer}, but their application to 3D medical imaging is challenging due to high computational cost and the need to preserve fine-grained spatial details~\cite{DATR}. Combining CNNs and Transformers in hybrid models can leverage the strengths of both, addressing the challenges of 3D landmark detection.

To address these challenges, we propose a hybrid framework, \textbf{H3DE-Net}, which combines the strengths of CNNs and Transformers for robust and accurate 3D landmark detection. CNNs are employed for efficient local feature extraction and multi-scale representation, while the Transformer module, equipped with a Volumetric Bi‑Routing Attention (V‑BRA) mechanism, enhances global context modeling. Additionally, a Super-Resolution Block (SRB) is introduced to recover fine-grained spatial details, improving landmark localization in low-resolution regions. A Feature Fusion Module (FFM) further integrates multi-scale features, effectively capturing both global and local dependencies. This synergistic design allows H3DE-Net to overcome the limitations of standalone architectures, achieving precise and reliable landmark detection even in challenging scenarios.

Our main contributions are as follows:
\begin{itemize}
    \item We design a Volumetric Bi‑Routing Attention (V‑BRA) model for 3D landmark detection, combining CNNs for local feature extraction with an extended BiFormer for global context, achieving state-of-the-art results in challenging scenarios.
    \item We introduce a Feature Fusion Module (FFM) for multi-scale integration and a Super-Resolution Block (SRB) to enhance precision and robustness.
    \item Extensive experiments show that our method outperforms existing baselines, especially in complex anatomical structures and missing landmarks.
\end{itemize}

\section{Method}\label{Sec.3}

\subsection{Problem Definition}
This study focuses on detecting anatomical landmarks in head 3D-CT scans, where the goal is to predict the 3D coordinates of \( N \) landmarks \( L = \{l_1, l_2, \ldots, l_N\} \), with each landmark \( l_i = (x_i, y_i, z_i) \).

We formulate the detection task as a heatmap regression problem, where each landmark \( l_i \) is represented by a Gaussian distribution. The heatmap for each landmark is defined as:
\begin{equation}
H_i(x, y, z) = \exp\left(-\frac{(x - x_i)^2 + (y - y_i)^2 + (z - z_i)^2}{2\sigma^2}\right),
\end{equation}
and the final heatmap is the sum of all individual heatmaps: \( H = \sum_{i=1}^{N} H_i \).

During inference, the predicted coordinate \( \hat{l}_i \) is obtained by finding the peak of the predicted heatmap:
\begin{equation}
\hat{l}_i = (\hat{x}_i, \hat{y}_i, \hat{z}_i) = \arg\max_{(x, y, z)} H(x, y, z).
\end{equation}

\begin{figure}[htbp]
\centering
\includegraphics[width=0.8\linewidth,trim={0cm 0cm 0cm 0cm},clip]{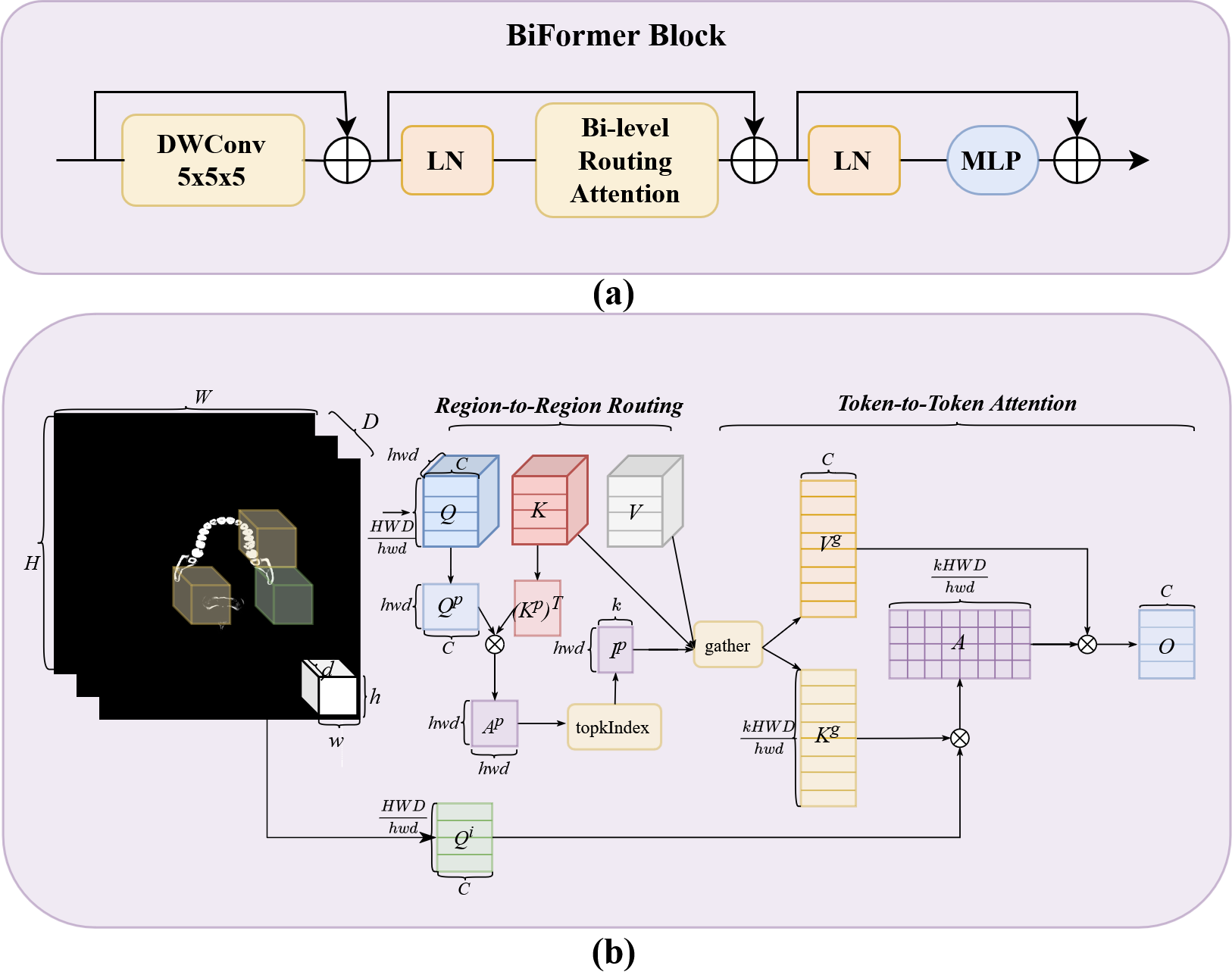}
\caption{(a) Architecture of the original Biformer Block. (b) Illustration of region-to-region routing and token-to-token attention. Our method uses sparsity by selecting key-value pairs from the top-$k$ most relevant windows, avoiding unnecessary calculations.}
\label{fig:biformer}
\end{figure}

\subsection{Volumetric Bi‑Routing Attention (V‑BRA) Design}
\begin{figure*}[htbp]
\centering
\includegraphics[width=0.8\linewidth,trim={0cm 0cm 0cm 0cm},clip]{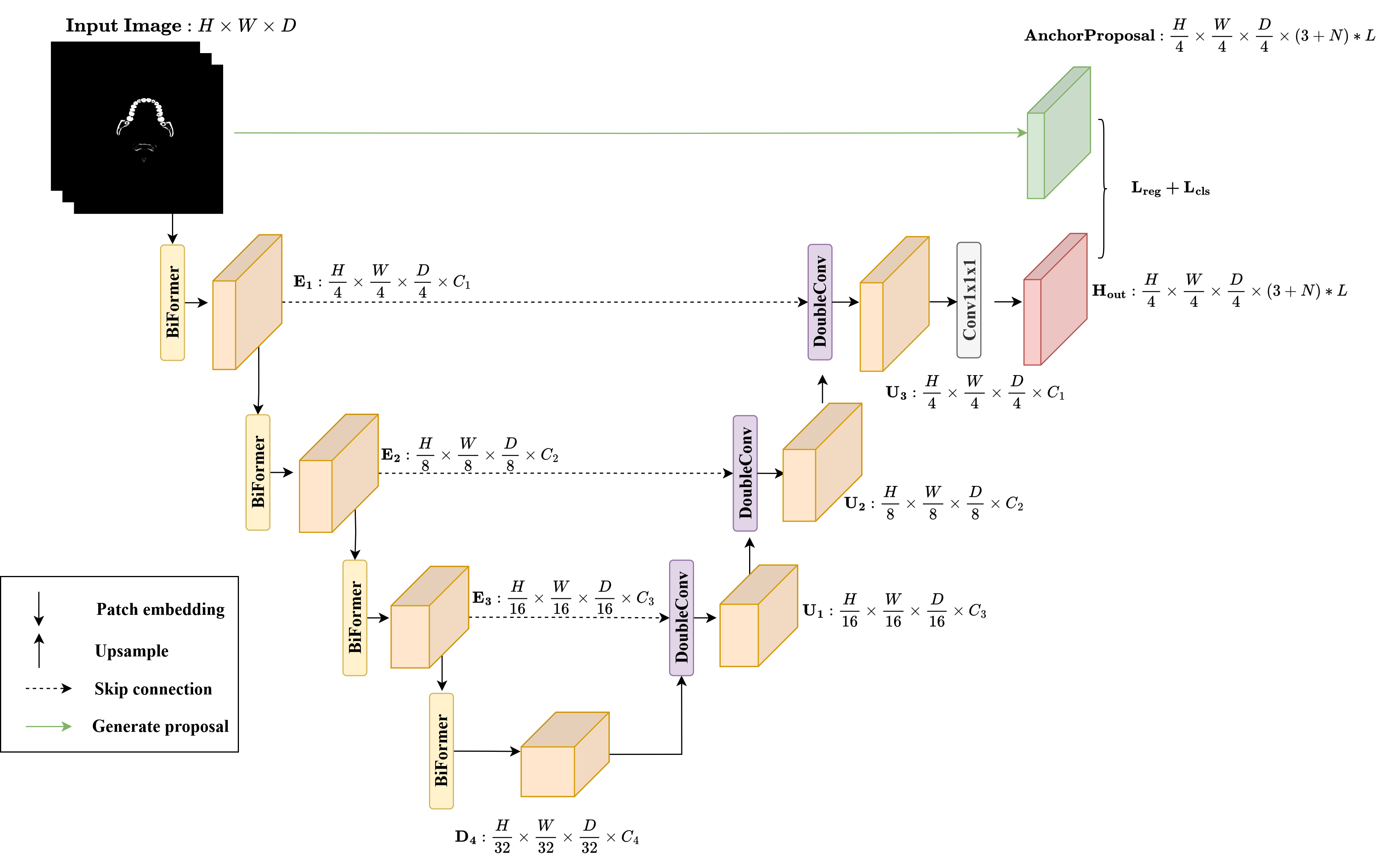}
\caption{Overview of the proposed Hybrid-3D Network (H3DE-Net): Anchor-Based Architectures.}
\label{fig: network2}
\end{figure*}

The Volumetric Bi‑Routing Attention (V‑BRA) Module is the core of H3DE-Net, designed to efficiently extract global and local features from 3D data. By extending the BiFormer module~\cite{biformer} to 3D, it introduces a bi-level routing attention mechanism that reduces computational complexity while maintaining the ability to model long-range dependencies and local details. The module consists of:

\begin{itemize}
    \item Depthwise Convolution (DWConv): Captures local spatial features.
    \item Volumetric Bi‑Routing Attention (V-BRA): Models global and local dependencies, including Region-to-Region Routing and Token-to-Token Attention.
    \item Layer Normalization (LN) and Multi-Layer Perceptron (MLP): Enhance feature distributions and representations.
\end{itemize}

The V-BRA mechanism operates in two stages:

\paragraph{Region-to-Region Routing}  
The input feature \( F \in \mathbb{R}^{H \times W \times D \times C} \) is divided into regions. Coarse-grained attention is computed using query \( Q^p \) and key \( K^p \), defined as:
\begin{equation}
\text{Attention}_{\text{coarse}}(Q^p, K^p) = \text{Softmax}\left( \frac{Q^p (K^p)^\top}{\sqrt{d_k}} \right),
\end{equation}
where \( Q^p, K^p \in \mathbb{R}^{(h \times w \times d) \times C} \), and \( d_k \) is the attention head dimension. Coarse-grained routing selects high-relevance regions through a Top-k mechanism, reducing computational burden.

\paragraph{Token-to-Token Attention}  
Within selected regions, fine-grained attention is computed to capture detailed features:
\begin{equation}
\text{Attention}_{\text{fine}}(Q^*, K^*, V^*) = \text{Concat}(\text{head}_1, \ldots, \text{head}_h) W_O,
\end{equation}
where \( Q^*, K^*, V^* \in \mathbb{R}^{k \times d_k} \) are features of the selected tokens, and \( W_O \) is a linear projection matrix. The aggregated attention produces the output features \( O \in \mathbb{R}^{H \times W \times D \times C} \).




\subsection{Designed Architectures}
To accommodate different task requirements(We cover this in detail in Sec.~\ref{sec4}), we design two architectures: \textbf{Anchor-Free Architecture} and \textbf{Anchor-Based Architecture}. Both architectures follow a unified encoder-decoder framework, employ the Volumetric Bi‑Routing Attention (V‑BRA) module for multi-scale feature extraction, and integrate additional components such as the \textbf{Feature Fusion Module (FFM)}. These components further enhance the model's capability to aggregate multi-scale information and recover spatial details, which is critical for accurate landmark localization.

\subsubsection{Anchor-Based Architecture}
\begin{figure}[htbp]
\centering
\includegraphics[width=0.8\linewidth,trim={0cm 0cm 0cm 0cm},clip]{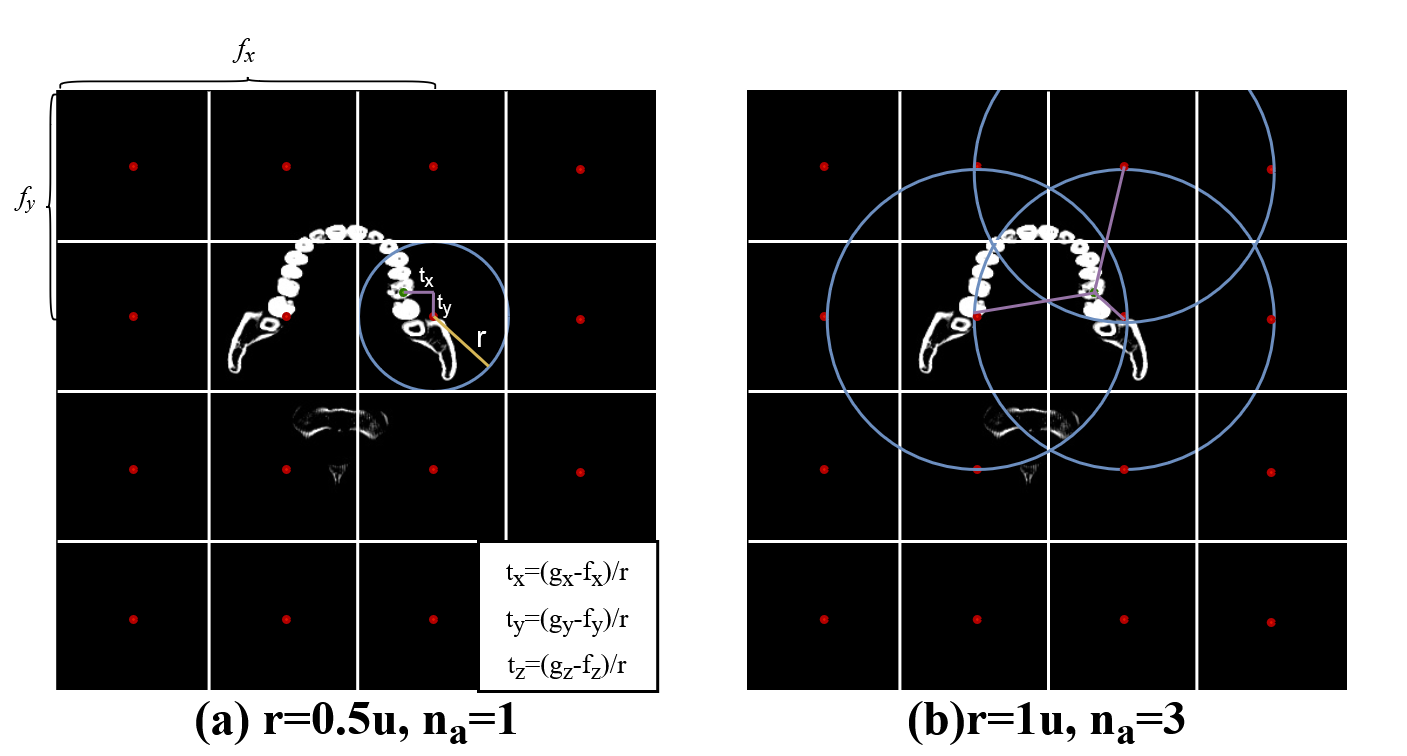}
\caption{(a) shows a single-scale anchor design where \( r = 0.5u \) and the number of anchors \( n_a = 1 \), while (b) shows a multi-scale anchor design where \( r = 1u \) and \( n_a = 3 \). The multi-scale design enhances the coverage of regions surrounding partially missing landmarks, making the model more robust in handling irregular landmark distributions and incomplete data.}
\label{fig:anchor}
\end{figure}

Figure~\ref{fig:anchor} illustrates the anchor design, which addresses the issue of missing landmarks in the dataset. Anchors are uniformly distributed in a grid layout across the 3D feature space, with fixed intervals and radii \( r \). Each anchor serves as a candidate for predicting landmark positions through offset regression and existence probability estimation. The offset regression is parameterized as:
\begin{equation}
t_x = \frac{g_x - f_x}{r}, \quad t_y = \frac{g_y - f_y}{r}, \quad t_z = \frac{g_z - f_z}{r},
\end{equation}
where \( (g_x, g_y, g_z) \) are the ground-truth coordinates of the landmark, \( (f_x, f_y, f_z) \) are the anchor center coordinates, and \( r \) is the anchor radius. The two subfigures in Figure~\ref{fig:anchor} demonstrate different anchor configurations. 

The Anchor-Based architecture introduces an anchor proposal mechanism that divides the 3D space into candidate regions and predicts the offsets and probabilities for each anchor. As shown in Figure~\ref{fig: network2}, the architecture consists of an encoder and a decoder. The encoder extracts multi-scale features from the input 3D image \( H \times W \times D \) using the Volumetric Bi‑Routing Attention (V‑BRA) module. Each encoder stage generates feature maps \( E_i \) with progressively reduced spatial dimensions:
\begin{equation}
E_i = \text{V-BRA}(E_{i-1}), \quad i = 1, 2, \ldots, 4,
\end{equation}
where \( E_i \in \mathbb{R}^{\frac{H}{2^i} \times \frac{W}{2^i} \times \frac{D}{2^i} \times C_i} \). 

The decoder reconstructs spatial resolution through upsampling layers and skip connections. The final decoder stage outputs \( U_1 \), which is used to generate two prediction branches:
1) An offset regression branch that predicts the offsets \( \Delta l \) of each anchor relative to the ground-truth landmark positions:
\begin{equation}
\Delta l = f_{\text{offset}}(U_1), \quad \Delta l \in \mathbb{R}^{\frac{H}{4} \times \frac{W}{4} \times \frac{D}{4} \times (3 \cdot L)},
\end{equation}
where \( L \) is the number of landmarks, and the factor of 3 corresponds to the x, y, and z coordinates in 3D space.
2) A probability prediction branch that estimates the existence probability \( p \) of a landmark for each anchor. The probabilities are normalized using the sigmoid activation function \( \sigma \):
\begin{equation}
p = \sigma(f_{\text{prob}}(U_1)), \quad p \in \mathbb{R}^{\frac{H}{4} \times \frac{W}{4} \times \frac{D}{4} \times L}.
\end{equation}


By focusing on candidate regions, the Anchor-Based architecture captures landmark-specific local features more accurately, making it particularly effective for scenarios with irregularly distributed or missing landmarks. 

\subsubsection{Anchor-Free Architecture}
The Anchor-Free architecture directly generates a 3D heatmap \( H_{\text{out}} \in \mathbb{R}^{H \times W \times D \times L} \), where \( H, W, D \) represent the spatial dimensions of the input image, and \( L \) is the total number of landmarks. The network models landmarks globally through encoding, decoding, and feature fusion, following the structure shown in Fig. 2, but without the anchor mechanism.

Given a 3D input image \( I \in \mathbb{R}^{H \times W \times D} \), an initial convolution layer extracts features:
\begin{equation}
E_0 = \text{Conv}_{3\times3}(I), \quad E_0 \in \mathbb{R}^{H \times W \times D \times C_0}.
\end{equation}

The encoder extracts multi-scale features \( E_i \) (\( i = 1, 2, \ldots, 5 \)) through successive Volumetric Bi‑Routing Attention (V‑BRA) modules:
\begin{equation}
E_i = \text{V-BRA}(E_{i-1}), \quad i = 1, 2, \ldots, 4.
\end{equation}

The decoder restores spatial resolution through upsampling layers \( U_j \) (\( j = 0, 1, 2, 3 \)), combined with skip connections from the encoder($U_0$ is equivalent to $D_4$ in Figure~\ref{fig: network2}):
\begin{equation}
U_j = \text{Fusion}(U_{j-1}, E_{4-j}), \quad j = 1, 2, 3.
\end{equation}

The final heatmap \( H_{\text{out}} \) is produced via a \( 1 \times 1 \times 1 \) convolution:
\begin{equation}
H_{\text{out}} = \text{Conv}_{1\times1\times1}(U_3), \quad H_{\text{out}} \in \mathbb{R}^{H \times W \times D \times L}.
\end{equation}

\subsection{Loss Function}
The network is optimized using a hybrid loss function tailored to each architecture. For the Anchor-free architecture, the loss is based solely on heatmap regression, while for the Anchor-based architecture, additional anchor-related terms are included.

The heatmap regression loss is defined as:

\begin{align}
\mathcal{L}_{\text{heatmap}} = \frac{1}{HWDN} 
\sum_{n=1}^N & \sum_{h=1}^H \sum_{w=1}^W \sum_{d=1}^D \notag \\
& \left( H(h, w, d, n) - G(h, w, d, n) \right)^2,
\end{align}

where \( H(h, w, d, n) \) and \( G(h, w, d, n) \) are the predicted and ground truth heatmap values for the \( n \)-th landmark.

For the Anchor-based architecture, the anchor regression and classification losses are:

\begin{equation}
\mathcal{L}_{\text{reg}} = \frac{1}{N} \sum_{i=1}^N \| (\Delta x_i, \Delta y_i, \Delta z_i) - (\hat{\Delta x}_i, \hat{\Delta y}_i, \hat{\Delta z}_i) \|^2,
\end{equation}

\begin{equation}
\mathcal{L}_{\text{cls}} = - \frac{1}{N} \sum_{i=1}^N \left[ p_i \log \hat{p}_i + (1 - p_i) \log (1 - \hat{p}_i) \right],
\end{equation}

where \( (\Delta x_i, \Delta y_i, \Delta z_i) \) and \( (\hat{\Delta x}_i, \hat{\Delta y}_i, \hat{\Delta z}_i) \) are the ground truth and predicted offsets for the \( i \)-th anchor, and \( p_i \) and \( \hat{p}_i \) are the ground truth and predicted existence probabilities.

For the Anchor-free architecture, the total loss is simply:

\begin{equation}
\mathcal{L}_{\text{total}} = \mathcal{L}_{\text{heatmap}}.
\end{equation}

For the Anchor-based architecture, the total loss combines all components:

\begin{equation}
\mathcal{L}_{\text{total}} = \lambda_{\text{reg}} \mathcal{L}_{\text{reg}} + \lambda_{\text{cls}} \mathcal{L}_{\text{cls}} + \lambda_{\text{heatmap}} \mathcal{L}_{\text{heatmap}}.
\end{equation}

Here, \( \lambda_{\text{reg}} \) and \( \lambda_{\text{cls}} \) are hyperparameters controlling the contributions of the anchor-related losses.

\section{Experiment}
\subsection{Dataset Preprocessing and Evaluation Metric}\label{sec4}
To validate the effectiveness and robustness, we compare with several widely used baseline methods in medical image analysis: UNet3D~\cite{3dunet}, VNet~\cite{3dvnet}, ResUNet3D~\cite{3dresunet}, HRNet~\cite{HRnet}, SRPose~\cite{SRpose}, MTL~\cite{Zhang2019ContextguidedFC}, Two-Stage~\cite{HE202115}, and SRLD-UNet \& SR-UNet~\cite{3D-CNN2-sr}.

Dataset: We evaluate H3DE-Net on a publicly available 3D skull landmark detection dataset~\cite{3D-CNN1-anchor} with 658 CT scans (458 training, 100 validation, 100 testing), annotated with 14 mandibular molar landmarks. The dataset has two subsets: Complete Cases, where all landmarks are annotated (283 training, 56 validation, 60 testing), and Incomplete Cases, where some landmarks are missing due to occlusion or artifacts (175 training, 44 validation, 40 testing), simulating real-world conditions.

Preprocessing: We standardize inputs, reduce computational cost, and improve model robustness through:
\begin{itemize}
    \item Intensity normalization: Raw CT voxel values are normalized to the range $[0, 255]$.
    \item Random cropping: Each CT volume is cropped to a fixed size of $128 \times 128 \times 64$ voxels.
\end{itemize}

Evaluation Metric: We use mean radial error (MRE) and successful detection rate (SDR) as evaluation metrics. MRE represents the average Euclidean distance between predicted and true values, and SDR measures detection accuracy at different distances (2mm, 2.5mm, 3mm, and 4mm).

\begin{table*}[htbp]
\centering
\footnotesize 
\caption{The experimental results for VNet, UNet3D, ResUNet3D, and H3DE-Net on the \texttt{All}, \texttt{Complete}, and \texttt{Incomplete} datasets. “$\checkmark$” indicates the use of an anchor.}
\label{tab:h3de_results}
\setlength{\tabcolsep}{7pt} 
\resizebox{0.8\linewidth}{!}{ 
\begin{tabular}{|l|c|c|c|c|c|c|c|}
\hline
\multirow{2}{*}{\textbf{Network}} & \multirow{2}{*}{\textbf{Dataset}} & \multirow{2}{*}{\textbf{Anchor}} & \multirow{2}{*}{\textbf{MRE$\pm$Std (mm)}} & \multicolumn{4}{c|}{\textbf{SDR (\%)}} \\
\cline{5-8}
 & & & & \textbf{2mm} & \textbf{2.5mm} & \textbf{3mm} & \textbf{4mm} \\
\hline
VNet~\cite{3dvnet}     & All         & $\checkmark$ & 1.82 $\pm$ 0.73 & 71.72 & 83.50 & 90.07 & 94.95 \\
                       & Complete    & $\times$     & 1.99 $\pm$ 0.71 & 67.86 & 78.31 & 85.20 & 91.20 \\
                       & Incomplete  & $\checkmark$ & 2.59 $\pm$ 1.82 & 49.19 & 64.05 & 75.68 & 90.54 \\
\hline
UNet3D~\cite{3dunet}   & All         & $\checkmark$ & 1.77 $\pm$ 0.87 & 73.97 & 85.21 & 91.98 & 96.45 \\
                       & Complete    & $\times$     & 1.90 $\pm$ 0.65 & 65.94 & 77.81 & 86.99 & 93.54 \\
                       & Incomplete  & $\checkmark$ & 2.26 $\pm$ 1.26 & 61.89 & 74.86 & 82.43 & 91.89 \\
\hline
ResUNet3D~\cite{3dresunet} & All    & $\checkmark$ & 1.70 $\pm$ 0.72 & \textbf{76.43} & 86.45 & 90.91 & 95.20 \\
                           & Complete & $\times$    & 1.96 $\pm$ 0.82 & 70.03 & 79.97 & 86.10 & 92.73 \\
                           & Incomplete & $\checkmark$ & 2.29 $\pm$ 1.44 & 58.65 & 72.79 & 81.08 & 89.46 \\
\hline
\textbf{H3DE-Net (Ours)} & All         & $\checkmark$ & \textbf{1.67 $\pm$ 0.64} & 76.03 & \textbf{86.50} & \textbf{92.31} & \textbf{96.69} \\
                         & Complete    & $\times$     & \textbf{1.68 $\pm$ 0.45} & \textbf{71.19} & \textbf{85.24} & \textbf{91.67} & \textbf{97.02} \\
                         & Incomplete  & $\checkmark$ & \textbf{2.07 $\pm$ 0.96} & \textbf{64.55} & \textbf{76.61} & \textbf{85.29} & \textbf{93.72} \\
\hline
\end{tabular}
}
\end{table*}

\subsection{Experimental Results}
Experiments are conducted on three dataset configurations: 1) \textit{complete dataset}, where all landmarks are annotated, requiring no Anchor-based Architecture; 2) \textit{incomplete dataset}, where some landmarks are missing due to occlusion or artifacts, requiring Anchor-based Architecture; 3) \textit{all cases dataset}, combining complete and incomplete cases to evaluate generalization, where Anchor-based Architecture is used.

Table \ref{tab:h3de_results} summarizes the quantitative results of H3DE-Net compared to baseline methods such as VNet, UNet3D, and ResUNet3D. On the \textit{complete dataset}, H3DE-Net achieves an MRE of $1.68 \pm 0.45$mm and an SDR of 97.02\% at a 4mm threshold, outperforming UNet3D, which has an MRE of $1.90 \pm 0.65$mm and an SDR of 93.54\%.

On the All Cases dataset, H3DE-Net achieves an MRE of $1.67 \pm 0.64$mm, outperforming the baselines. At the 2mm and 4mm thresholds, SDR reaches 76.03\% and 96.69\%, while ResUNet3D achieves 76.43\% and 95.20\%, respectively. These results demonstrate the model's ability to generalize across heterogeneous datasets, capturing anatomical variations without sacrificing accuracy.

Even on the challenging \textit{incomplete dataset}, where missing landmarks and imaging artifacts complicate localization, H3DE-Net shows robust performance, with an MRE of $2.07 \pm 0.96$mm and SDR (4mm) of 93.72\%, outperforming UNet3D (MRE: $2.26 \pm 1.26$mm, SDR (4mm): 91.89\%).

We also compare H3DE-Net with competitive methods such as HRNet-32, HRNet-48, SRPose, Two-Stage, MTL, SRLD-UNet, and SR-UNet. Table~\ref{tab:tab2} summarizes the results in terms of MRE and SDR at thresholds of 2mm, 2.5mm, 3mm, and 4mm.

\begin{table}[htbp]
\centering
\footnotesize 
\caption{The performance comparison between H3DE-Net and other SOTA methods. Results are extracted from~\cite{3D-CNN2-sr}, where ‘--’ indicates missing values.}
\label{tab:tab2}
\setlength{\tabcolsep}{5pt} 
\begin{tabular}{|l|c|c|c|c|c|}
\hline
\multirow{2}{*}{\textbf{Network}} & \multirow{2}{*}{\textbf{MRE$\pm$Std (mm)}} & \multicolumn{4}{c|}{\textbf{SDR (\%)}} \\
\cline{3-6}
 & & \textbf{2mm} & \textbf{2.5mm} & \textbf{3mm} & \textbf{4mm} \\
\hline
HRNet-32~\cite{HRnet}     & 2.64 $\pm$ 5.18  & 60.81 & --    & 82.27 & 90.22 \\
HRNet-48~\cite{HRnet}     & 2.77 $\pm$ 5.83  & 58.89 & --    & 82.05 & 90.16 \\
SRPose~\cite{SRpose}      & 2.58 $\pm$ 5.73  & 70.09 & --    & 87.27 & 92.46 \\
Two-Stage~\cite{HE202115} & 1.78 $\pm$ 0.81  & 68.93 & 82.48 & 88.68 & 95.37 \\
MTL~\cite{Zhang2019ContextguidedFC} & 1.91 $\pm$ 0.75  & 69.70 & 80.47 & 87.29 & 93.86 \\
SRLD-UNet~\cite{3D-CNN2-sr} & 2.40 $\pm$ 5.63 & 74.98 & --    & 90.39 & 94.45 \\
SR-UNet~\cite{3D-CNN2-sr}   & 2.01 $\pm$ 4.33 & \textbf{76.14} & -- & 92.02 & 95.84 \\
\textbf{H3DE-Net (Ours)}    & \textbf{1.67 $\pm$ 0.64} & 76.03 & \textbf{86.50} & \textbf{92.31} & \textbf{96.69} \\
\hline
\end{tabular}
\end{table}

\begin{figure*}[htbp]
\centering
\begin{minipage}{\textwidth}
  \centering
  \includegraphics[width=\linewidth,trim={0cm 0cm 0cm 0cm},clip]{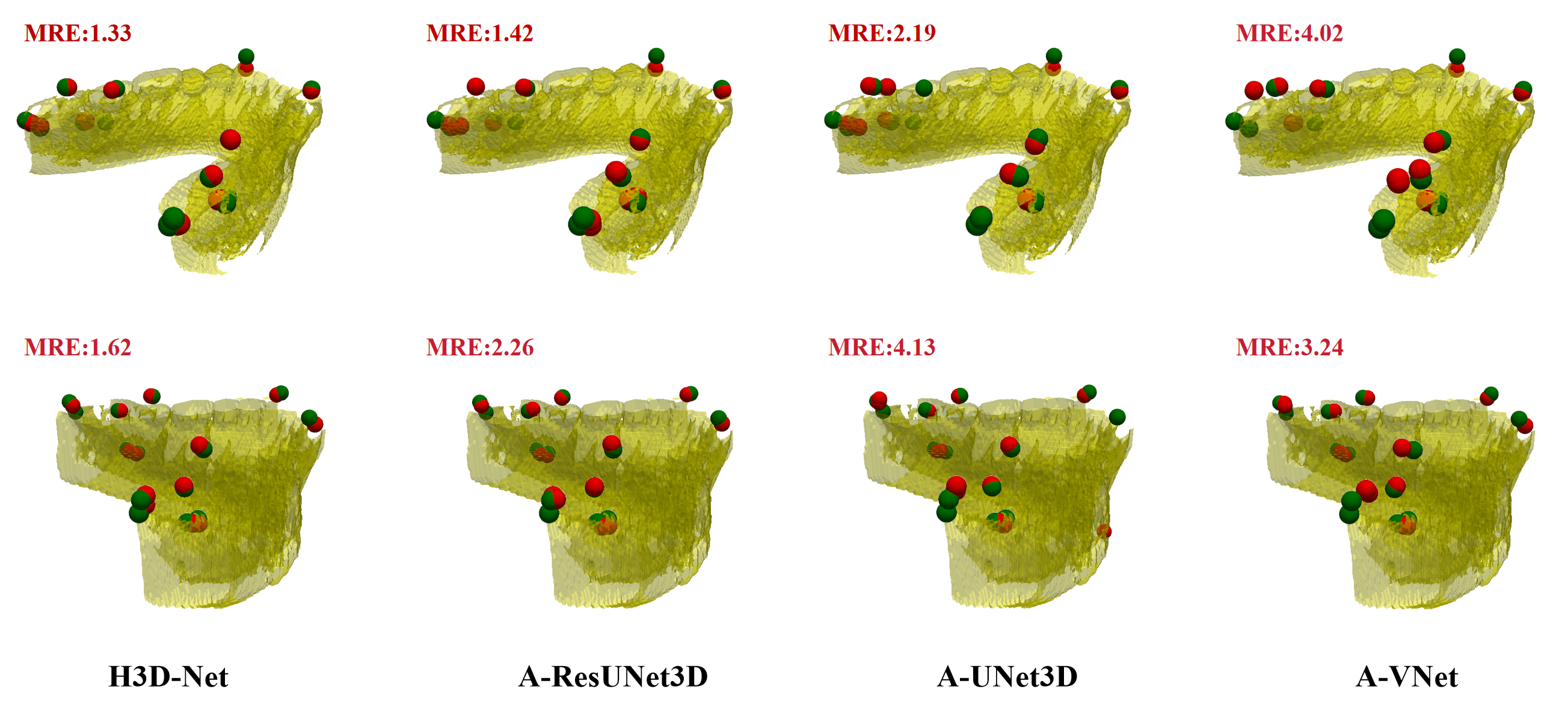}
  \caption{Landmark detection performance of H3DE-Net train and test on the all datasets. The prefix 'A-' represents the Anchor-Based method, and the way anchors are added is the same as shown in Fig.~\ref{fig: network2}.}
  \label{fig: p}
\end{minipage}
\end{figure*}


Among all comparison methods, H3DE-Net demonstrates superior performance across all metrics, with significant improvements in accuracy and robustness. Compared to HRNet-32 and HRNet-48, H3DE-Net reduces MRE by 36.74\% and 39.71\%, respectively, while improving SDR (4mm) by 6.47\% and 6.53\%. Though the Two-Stage method performs better in MRE (1.78 $\pm$ 0.81mm), it lags behind H3DE-Net in SDR (2mm) and SDR (3mm) by 7.10\% and 3.63\%, respectively, highlighting H3DE-Net's advantage in high-precision detection.

Compared to MTL, H3DE-Net reduces MRE by 12.57\% and improves SDR (4mm) by 2.83\%, confirming its strong generalization ability. Notably, compared to SRPose and SRLD-UNet, H3DE-Net significantly improves SDR at all thresholds, especially at 2mm, where it increases by 5.94\% and 1.59\%, respectively, showing its superiority in detecting small-range errors.

These results demonstrate that H3DE-Net's innovative architecture, which integrates multi-scale features and combines global and local information, achieves state-of-the-art performance by effectively handling complex data, missing, or noisy landmarks.

We visually analyze the experimental results, where ground truth landmarks are shown in green and predicted landmarks in red. Figure~\ref{fig: p} presents the visualization of three Anchor-Based methods and H3DE-Net using all training data. H3DE-Net demonstrates superior localization accuracy, particularly in anatomically complex regions like the nasal cavity and skull base, where baseline methods often show significant deviations.

Overall, the results demonstrate the consistent superiority of H3DE-Net across all dataset configurations. The proposed method not only outperforms baseline models under ideal conditions but also excels in real-world scenarios where incomplete or heterogeneous data is prevalent. These findings validate the robustness and versatility of the proposed framework, establishing it as a reliable tool for landmark localization tasks in medical imaging.




\section{Conclusion}
This paper introduces H3DE-Net, a hybrid framework for 3D landmark detection in medical images. By integrating CNNs and Transformers, it addresses key challenges of volumetric data, including sparse landmark distribution, complex anatomy, and multi-scale dependencies. The CNN backbone captures local and multi-scale features, while the Volumetric Bi‑Routing Attention (V‑BRA) module efficiently models global context via bi-level routing attention. A feature fusion module further enhances robustness and precision. Experiments on a public dataset show that H3DE-Net outperforms existing methods, achieving state-of-the-art accuracy, particularly in scenarios with missing landmarks or anatomical variations.

\bibliographystyle{IEEEtran}
\bibliography{reference}

\end{document}